\documentclass{article}

\usepackage[preprint, final]{neurips_2025}
\usepackage{subcaption}
\usepackage{float}

\usepackage[utf8]{inputenc} 
\usepackage[T1]{fontenc}    
\usepackage{hyperref}       
\usepackage{url}            
\usepackage{booktabs}       
\usepackage{amsfonts}       
\usepackage{nicefrac}       
\usepackage{microtype}      
\usepackage{xcolor}         
\usepackage{multirow}
\usepackage{graphicx}
\usepackage{amsmath}

\title{MoRE-GNN: Multi-omics Data Integration with a Heterogeneous Graph Autoencoder}

\author{%
Zhiyu Wang\thanks{Those authors contributed equally.} \\
University of Cambridge \\
\texttt{zw471@cam.ac.uk} \And
Sonia Koszut \footnotemark[1]\\
University of Cambridge \\
\texttt{smk79@cam.ac.uk} \And
Pietro Li\`o \\
University of Cambridge \\
\texttt{pl219@cam.ac.uk} \And
Francesco Ceccarelli \thanks{Corresponding author.} \\
University of Cambridge \\
\texttt{fc485@cam.ac.uk} 
}

\begin{document}

\maketitle

\begin{abstract}
The integration of multi-omics single-cell data remains challenging due to high-dimensionality and complex inter-modality relationships. To address this, we introduce MoRE-GNN (Multi-omics Relational Edge Graph Neural Network), a heterogeneous graph autoencoder that combines graph convolution and attention mechanisms to dynamically construct relational graphs directly from data. Evaluations on six publicly available datasets demonstrate that MoRE-GNN captures biologically meaningful relationships and outperforms existing methods, particularly in settings with strong inter-modality correlations. Furthermore, the learned representations allow for accurate downstream cross-modal predictions. While performance may vary with dataset complexity, MoRE-GNN offers an adaptive, scalable and interpretable framework for advancing multi-omics integration. 
The code is made available at \url{https://github.com/ZW471/MultiOmicsIntegration}.
\end{abstract}

\section{Introduction}
\label{sc: introduction}



Multi-omics data integration has emerged as a crucial challenge in bioinformatics. By combining information from diverse molecular modalities---such as genomics, transcriptomics, epigenomics, and proteomics---researchers can obtain a more holistic view of biological systems and disease mechanisms~\cite{baiao2025technical}. Integrating multiple omics layers can reveal relationships invisible to single-omics analysis, advancing precision medicine and our understanding of complex cellular processes~\cite{baiao2025technical}. However, this integration is non-trivial due to high-dimensional, heterogeneous data with varying scales, noise characteristics, and missing values~\cite{hasin2017multi}. Adding to this complexity, each modality only captures a slice of the biology~\cite{LOVINO2022494}, and cross-modal relationships may be nonlinear or context-dependent.

Key challenges include accounting for batch effects, lack of feature correspondences across modalities, and modeling interactions between different data types~\cite{vilanova2016multi}. Traditional statistical methods~\cite{Bersanelli2016, Jiang2023} assume linear relationships and require paired samples, limiting their ability to capture biological complexity. On the other hand, advanced machine learning approaches must contend with scalability issues and the risk of overfitting~\cite{cai2025integrating}. Recent progress spans from classical techniques to modern representation learning, including canonical correlation analyses, matrix factorization, and deep generative models~\cite{baiao2025technical}. Methods like Similarity Network Fusion~\cite{Wang2014} and Multi-Omics Factor Analysis~\cite{https://doi.org/10.15252/msb.20178124} partially capture nonlinear relationships, but often fall short in modeling complex feature-level interactions.

Graph Neural Networks (GNNs)~\cite{10670406, Murgod2024} have emerged as a promising approach for multi-omics integration by representing data as graphs and naturally capturing biological entity interactions. Recent models like GLUE~\cite{cao2022multi}, scMoGNN~\cite{wen2022graph}, and scMI~\cite{cai2025integrating} leverage GNNs for single-cell multi-omics integration. These methods excel at modeling nonlinear relationships by explicitly considering interactions among cells, genes, and proteins, enabling information propagation across complementary biological views. However, existing approaches come with several limitations: GLUE relies on fixed biological priors~\cite{cao2022multi}, scMoGNN uses computationally expensive fully connected graphs~\cite{wen2022graph}, and scMI employs time-intensive random walks~\cite{cai2025integrating}. While heterogeneous graph approaches with relational message passing have shown success in protein structure analysis~\cite{Zhang2022ProteinPretraining}, these methods exploit 3D spatial relationships and sequential constraints that are not directly applicable to single-cell multi-omics data, which lack such inherent structural organization. There remains a need for methods that dynamically construct graphs from data-driven insights while maintaining computational scalability and biological interpretability.

\begin{figure}
    \centering
    \includegraphics[width=1\linewidth]{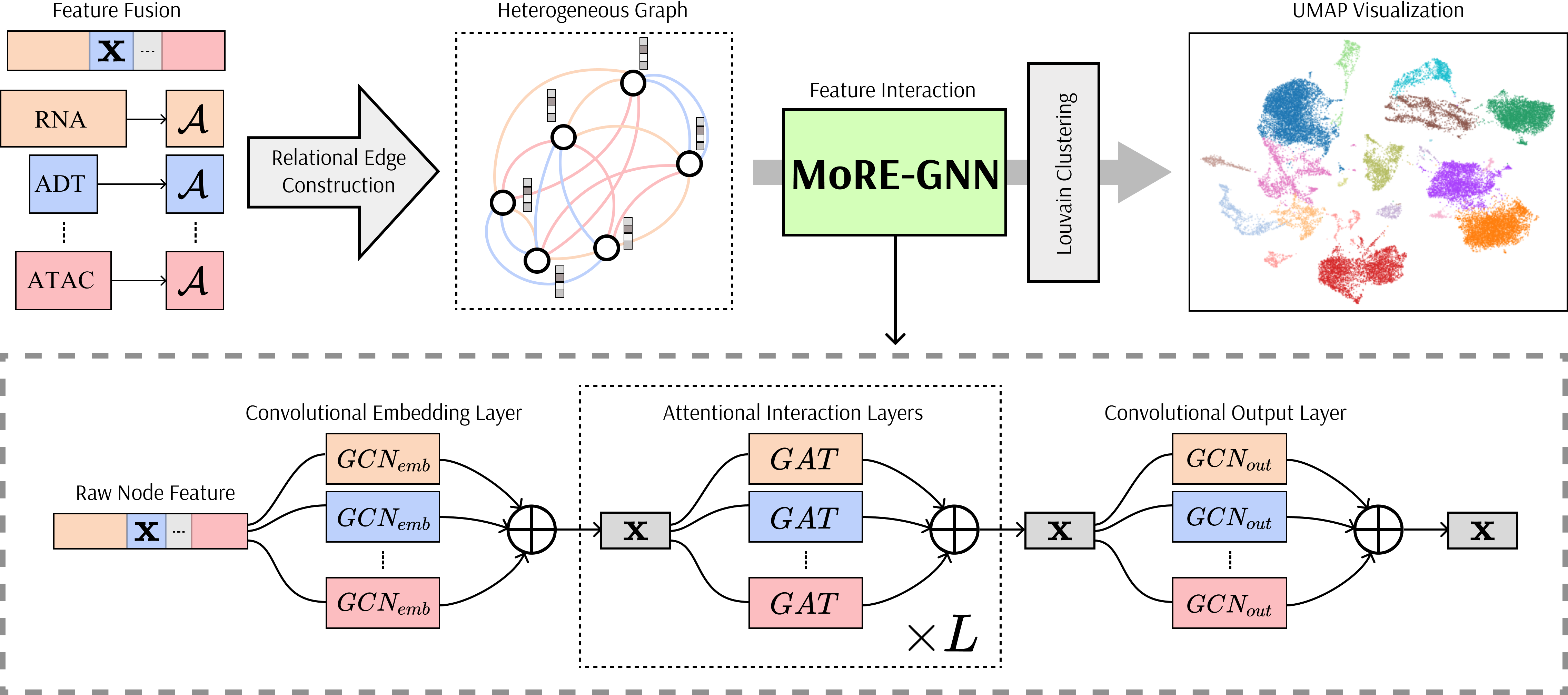}
    \caption{MoRE-GNN architecture for multi-omics single-cell integration. Multi-omics features (RNA, ATAC, ADT) are used to dynamically construct modality-specific adjacency matrices, resulting in a heterogeneous graph. MoRE-GNN processes this graph through: (1) convolutional embedding layers with modality-specific channels, (2) $L$ attentional interaction layers for cross-modal feature integration, and (3) a convolutional output layer. Final embeddings are clustered using Louvain clustering \cite{Blondel_2008} and reduced in dimensionality (UMAP) to reveal cell type structure.}
    \label{fig:visual_abstract}
\end{figure}

To address these challenges, we introduce Multi-omics Relational Edge Graph Neural Network (MoRE-GNN), a novel heterogeneous graph autoencoder framework for single-cell multi-omics integration. As illustrated in Figure~\ref{fig:visual_abstract}, MoRE-GNN models each cell as nodes within a dynamically constructed relational graph, where modality-specific adjacency matrices are derived entirely from data-driven similarity rather than predefined biological knowledge. Unlike existing heterogeneous graph methods that rely on fixed structural relationships (e.g., protein spatial arrangements), our approach constructs relational edges based on similarity within each omics modality, resulting in a flexible framework that adapts to the unique characteristics of the single-cell data at hand. Our method employs a combination of Graph Convolutional Networks (GCNs)~\cite{kipf2017semisupervisedclassificationgraphconvolutional} and attention mechanisms (GATv2)~\cite{brody2022attentivegraphattentionnetworks, Velickovic2017GraphNetworks}, capturing complex nonlinear interactions across multiple omics layers while maintaining computational efficiency for large-scale datasets. 

Our contributions are threefold: 
\begin{itemize}
    \item we introduce MoRE-GNN, a novel graph-based multi-omics integration method that dynamically constructs relational graphs purely from data, removing reliance on predefined biological priors and improving adaptability to diverse datasets; 
    \item we enhance the interpretability of multi-omics integration by modeling explicit cell-cell relationships and feature interactions, offering insights into cross-modal biological structure;
    \item we demonstrate that MoRE-GNN is able to learn biologically meaningful representations and excels in cross-modal prediction tasks, highlighting its potential for aiding and improving downstream analysis in single-cell multi-omics studies.
\end{itemize}

\section{Related Work}\label{sc: related-work}


\paragraph{Non-Parameterised Methods}
Early multi-omics integration efforts relied on statistical and matrix decomposition methods such as Canonical Correlation Analysis (CCA)~\cite{Jiang2023} and Partial Least Squares (PLS)~\cite{Bersanelli2016} to project high-dimensional omics data into shared latent spaces \cite{baiao2025technical}. More recently, methods like MOJITOO \cite{cheng2022mojitoo} introduced hybrid approaches combining linear projections with graph-based smoothing. While computationally efficient and interpretable, these techniques are fundamentally limited by their reliance on linear assumptions, making them less effective in capturing complex biological interactions.

\paragraph{Machine Learning-Based Methods}
To address linear limitations, ML-based approaches leverage dimensionality reduction, representation learning, and probabilistic modeling to derive shared embeddings. Classical techniques like Similarity Network Fusion (SNF) \cite{Wang2014} perform multi-view clustering by iteratively refining sample similarity graphs, while Multi-Omics Factor Analysis (MOFA) ~\cite{https://doi.org/10.15252/msb.20178124} incorporates latent factor models for cross-modal variation. Deep learning methods such as GLUE \cite{cao2022multi} apply variational autoencoders with graph-based priors, but rely heavily on predefined biological knowledge, limiting the discovery of novel relationships.

\paragraph{Graph Neural Network-Based Approaches}
Graph-based models explicitly encode relationships between biological entities for multi-omics integration. scMoGNN \cite{wen2022graph} models cells and molecular features as nodes in a bipartite graph using GCNs, but requires fully connected graphs, posing scalability challenges. scMI \cite{cai2025integrating} employs inter-type attention mechanisms with modality-specific graphs for improved integration, but is based on predefined structures that limit adaptability. 
MOGONET \cite{wang2021mogonet} utilizes multi-view graph convolutional networks with cross-modal attention to integrate omics data, though it relies on a fully supervised pipeline, requiring labeled datasets for training.

\section{Methodology}\label{sc: methodology}

MoRE-GNN (Figure~\ref{fig:visual_abstract}) operates in three stages:  (1) relational edges are constructed using cosine similarity for each modality, yielding graphs where similar cells are more strongly connected and multi-modal features are concatenated as node attributes;  (2) heterogeneous message passing is performed with GCN~\cite{kipf2017semisupervisedclassificationgraphconvolutional} and GATv2~\cite{brody2022attentivegraphattentionnetworks} layers to learn embeddings that capture modality-specific similarity relationships;  (3) the learned embeddings are projected into two dimensions using UMAP, and cell populations are identified with Louvain clustering.  The model is trained in a contrastive fashion, with modality-specific decoders predicting positive 
and negative edge links.

\subsection{Graph Construction}


Let $M$ denote the set of modalities. For each modality $m \in M$, we define a feature matrix $\mathbf{X}_m \in \mathbb{R}^{N \times d_m}$ where $N$ is the number of cells in the dataset and $d_m$ is the number of molecular features measured in modality $m$. The $i$-th row of $\mathbf{X}_m$ corresponds to the same cell across all modalities, i.e., rows are aligned by cell identity. For each modality $m$, we first calculate its similarity matrix:
\begin{align}
\label{eq: cosine-sim}
    S_m = \frac{\mathbf{x}_m \cdot \mathbf{x}_m}{\|\mathbf{x}_m\|_2^2 } \in \mathbb{R}^{N \times N}  .
\end{align}
The relational adjacency matrices $\{\mathcal{A}_m\}_{m \in M}$ are constructed from the similarity matrices 
$\{S_m\}_{m \in M}$ by retaining only the top $K$ entries in each row. Specifically, for each modality $m \in M$,
\begin{align}
\label{eq:rel-adj}
    \mathcal{A}_m(i,j) = 
    \begin{cases}
        1, & \text{if $S_{m,i,j}$ is among the top $K$ values in row $S_{m,i}$}, \\
        0, & \text{otherwise},
    \end{cases}
\end{align}
where $S_{m,i}$ denotes the $i^{\text{th}}$ row of $S_m$. Having obtained all the adjacency matrices, we build a graph $\mathcal{G}$ to represent all the cells in the dataset and their similarity relations: $ \mathcal{G} = (\mathcal{N}, \mathbf{X}, \{\mathcal{E}_m\}_{m \in M}) $, where $\mathcal{N}$ is the set of nodes, $\mathbf{X} = \big\|_{m \in M}\mathbf{x}_m $ correspond to the concatenated node features, and $\{\mathcal{E}_m\}_{m \in M}$ is the set of all relational edges. To allow for computational scalability and operational efficiency, we sample local subgraphs $\{\mathcal{G}_i\}_{i \in [0, |B|)}$ centered on $B$ seed cells that, for each modality, includes $N_1$ immediate neighbors. Then, $N_2$ neighbors are sampled for each of those, capturing the cell’s primary and secondary multi-modal interactions. Collectively, these samples partition the full graph into manageable subgraphs that approximate the global structure, enabling efficient mini-batch training while preserving both individual cell features and their relational context.

\subsection{Model}
\paragraph{Encoder}

Our proposed encoder framework is composed of a GCN~\cite{kipf2017semisupervisedclassificationgraphconvolutional} embedding block and a GCN output block at the start and at the end of the MoRE-GNN model, embedding original features to a latent embedding space and mapping the latent embedding to an output space, respectively. In the middle of the two GCN blocks are $L$ layers of GATv2~\cite{brody2022attentivegraphattentionnetworks} blocks that integrate attention mechanism in the learning process. Our model begins with embedding the original node features to a fixed dimension with GCN modules defined as
\begin{align}
    \mathbf{H}^{\prime} = \sigma\left( \hat{\mathbf{D}}^{-\frac{1}{2}} \hat{\mathcal{A}} \hat{\mathbf{D}}^{-\frac{1}{2}} \mathbf{H} \mathbf{W} \right), \label{eq: gcn}
\end{align}
where $\hat{\mathbf{D}} = \mathbf{D} + \mathbf{I}$ and $\hat{\mathcal{A}} = \mathcal{A} + \mathbf{I}$ are the degree matrix and the adjacency matrix with self-loops, respectively. The outputs from the GCN modules for each modality are summed, batch normalized, and passed through a non-linearity function $\sigma$. Thereafter, $L$ layers of GATv2 modules---each followed by the same aggregation and normalization procedure---are applied, enabling each node 
to learn from its neighbors through an attention mechanism. The message passing of the GATv2 layers is defined in Eq.~\ref{eq: gat layers}, and the GATv2~\cite{brody2022attentivegraphattentionnetworks} layer itself is defined as
\begin{align}
    h_i^{\prime} &= \sum_{j \in \mathcal{N}(i)} \alpha_{ij} \mathbf{W} h_j\\
    \alpha_{ij} &= \frac{ \exp\left( \text{LeakyReLU}\left( \mathbf{a}^\top [\mathbf{W} h_i \, \Vert \, \mathbf{W} h_j] \right) \right) }{ \sum_{k \in \mathcal{N}(i)} \exp\left( \text{LeakyReLU}\left( \mathbf{a}^\top [\mathbf{W} h_i \, \Vert \, \mathbf{W} h_k] \right) \right) },
\end{align}
where $\mathbf{a}$ is a learnable attention head. Each attention head maps the node embeddings from $\mathbb{R}^d$ to $\mathbb{R}^{d/n}$, where $n$ is the number of attention heads. The outputs of all heads are then concatenated to form the new node embedding $h_i^{\prime} \in \mathbb{R}^d$. Finally, MoRE-GNN concludes its message passing layers with another block of GCNs without normalization or nonlinearity as shown in Eq.~\ref{eq: gcn out}.
\begin{align}
    h_i^{l+1} &= \sigma\left(BN\left(\sum_{m \in M}GAT_m(h_i^l, \mathcal{A}_m)\right)\right) &\in \mathbb{R}^{d} \label{eq: gat layers} \\
    h_i^{out} &= \sum_{m \in M}GCN_{out,m}(h_i^{L}, \mathcal{A}_m) &\in \mathbb{R}^{d_{out} \label{eq: gcn out}}
\end{align}

\paragraph{Decoder}
There are $M$ modality-specific decoders $\{D_m\}_{m \in M}$ in our framework, each designed to reconstruct relational structures within its respective modality. Each decoder $D_m$ is implemented as a modality-specific multilayer perceptron (MLP) followed by a sigmoid activation, and operates on the element-wise product of the output embeddings of two nodes $h_i^{out}$ and $h_j^{out}$. Formally,
\begin{align}
    p_{m,(i,j)} = \sigma\!\left(\text{MLP}_m(h_i^{out} \circ h_j^{out})\right), 
    \quad p_{m,(i,j)} \in (0,1), \label{eq:decoder}
\end{align}
where $\circ$ denotes the element-wise (Hadamard) product. The scalar probability $p_{m,(i,j)}$ quantifies the likelihood of an edge between nodes $i$ and $j$ in modality $m$. An edge is predicted to exist if $p_{m,(i,j)} > 0.5$ ($\hat{y}_{(i,j),m}=1$), and to be absent otherwise ($\hat{y}_{(i,j),m}=0$). This decoder design is deliberately lightweight, enabling efficient training while retaining the capacity to capture modality-specific relational patterns, 
thereby allowing the model to focus computational resources on learning meaningful and robust node representations.

\paragraph{Loss Function}

To train MoRE-GNN, we define the loss function in Eq.~\ref{eq: total loss} to be the aggregation of $|M|$ modality-specific reconstruction losses, defined in Eq.~\ref{eq: recon loss}, and a clustering loss, defined in Eq.~\ref{eq: cluster loss}. The clustering loss, $\mathcal{L}_{cluster}$, encourages the learned representations to form compact clusters in the embedding space. Specifically, for each sample $i$ in the dataset, the loss computes the minimum Euclidean distance to a set of cluster centers $C$, where $|C|$ is a hyperparameter defining the number of clusters. This loss pushes the embeddings of $\mathbf{x}_i$ closer to its nearest cluster center, promoting intra-cluster compactness and inter-cluster separability. On the other hand, the reconstruction loss $\mathcal{L}_{recon, m}$ adopts a contrastive learning objective for each modality $m$ , using the set $E_{pos}$ of positive edge pairs and the set $E_{neg}$ of negative edge pairs. For the former, the model maximizes the predicted similarity $\hat{y}_{(i,j), m}$, while for the latter, it minimizes it, encouraging the model to reconstruct observed modality-specific relational structures accurately. The overall loss function balances these two objectives through the weighting hyperparameter $\alpha$, combining representation structure through clustering and data fidelity through reconstruction, to guide the model in learning robust and semantically meaningful multi-modal embeddings.
\begin{align}
    \mathcal{L}_{total} &= \alpha \cdot \mathcal{L}_{cluster} + \sum_{m \in M} \mathcal{L}_{recon, m} \label{eq: total loss}\\
    \mathcal{L}_{cluster} &= \frac{1}{|N|}\sum_{i\in N} min\{\|\mathbf{x}_i - \mathbf{x}_c\|_2\}_{c \in C}  \label{eq: cluster loss}\\
    \mathcal{L}_{recon, m} &= - \sum_{(i,j) \in E_{pos}}\log(\hat{y}_{(i,j), m}) - \sum_{(i,j) \in E_{neg}} \log(1 - \hat{y}_{(i,j), m}) \label{eq: recon loss}
\end{align}

\section{Experiments}\label{sc: experiments}
\subsection{Experimental Setup}
\paragraph{Datasets}

We evaluate our integration method on six publicly available single-cell multi-omics datasets which include bi- and tri-modal measurements, spanning different modalities, tissue types, and complexity levels: \texttt{BM-CITE} \cite{Stuart2021Single-cellSignac}, \texttt{LUNG-CITE} \cite{Buus2020ImprovingAnalysis}, \texttt{PBMC-Multiome} \cite{cheng2022mojitoo}, \texttt{PBMC-TEA} \cite{Swanson2021SimultaneousTea-seq}, \texttt{PBMC-DOGMA} \cite{Mimitou2021ScalableCells}, and \texttt{Skin-SHARE} \cite{Ma2020ChromatinChromatin}. 
\texttt{BM-CITE} and \texttt{LUNG-CITE} are canonical CITE-seq datasets from human bone marrow mononuclear cells and human peripheral blood mononuclear cells from lung respectively, containing gene expression and surface proteins information. The human peripheral blood mononuclear cells (\texttt{PBMC-Multiome}) is a 10x Genomics multiome dataset capturing RNA and ATAC-seq. Unlike CITE-seq, ATAC features tend to be high-dimensional and more sparse. \texttt{PBMC-TEA} and \texttt{PBMC-DOGMA} are two tri-modal PBMC datasets that measure RNA, ATAC, and protein data. These datasets test the scalability of our method to scenarios involving more than two modalities. Finally, \texttt{Skin-SHARE} is a SHARE-seq dataset from mouse skin. This dataset is more challenging than the others as differentiation trajectories yield continuous structures, requiring integration methods to respect gradual transitions rather than discrete clusters.
These datasets span a wide spectrum of integration difficulties, evaluating our method's ability to (i) align paired modalities with strong anchors, (ii) overcome sparsity and scale differences, (iii) preserve continuous differentiation structure, and (iv) extend to tri-modal settings.

\paragraph{Training Details}
For all experiments, the proposed MoRE-GNN model was implemented with a hidden dimension of $512$ and two GATv2 layers each having $8$ attention heads. The learning rate was set to $10^{-4}$ with a linear warm-up and cosine decay schedule, using $3$ warm-up epochs and a decay frequency of $1$. For the clustering 
loss, the number of clusters was fixed at $20$ and the loss weight at $0.01$, deliberately chosen to balance the clustering and reconstruction objectives, as the clustering term typically exhibits larger magnitude. Data preparation employed a batch size of $B=256$, with 1-hop and 2-hop neighborhood sample 
sizes $N_1=N_2=5$. Optimization was performed with Adam~\cite{kingma2017adammethodstochasticoptimization} ($\beta_1=0.9$, $\beta_2=0.999$, $\epsilon=10^{-8}$) without weight decay. All experiments were 
run on a single NVIDIA GeForce RTX 4070 GPU (8GB memory) for up to $500$ epochs, with early stopping triggered if the total loss did not decrease by at least $0.001$ for more than $3$ consecutive epochs. Across datasets, convergence was reached within $42$--$177$ epochs, with training times ranging from approximately $2$ to $40$ minutes depending on input size. For clustering, Louvain was consistently applied with resolution $0.5$ across all models and 
datasets.


\subsection{Cell Type Clustering}

We evaluate MoRE-GNN's clustering performance using Adjusted Rand Index (ARI) 
and Normalized Mutual Information (NMI)~\cite{hubert1985comparing_ari,vinh2009information_nmi}. We benchmark these metrics against those produced by MOJITOO \cite{cheng2022mojitoo}---a parameter-free approach based on canonical correlation analysis---serving as our baseline method. We compare primarily against MOJITOO as it represents the current state-of-the-art scalable integration method that, like MoRE-GNN, constructs relationships purely from data without relying on predefined biological knowledge. Table~\ref{tab:merged_results} reports ARI and NMI for MOJITOO and MoRE-GNN across the six above-mentioned datasets.


\begin{table}[tb]
    \centering
    \caption{Comparison of MOJITOO's and MoRE-GNN's clustering performance across six datasets. Our model exhibits superior performance on datasets with strong inter-modal correlations, but underperforms on datasets with more complex or noisy signals.}
    \label{tab:merged_results}
    \resizebox{\textwidth}{!}{
    \begin{tabular}{llcccccc}
        \toprule
        \textbf{Metric} & \textbf{Model} & \textbf{BM-CITE} & \textbf{LUNG-CITE} & \textbf{PBMC-Multiome} & \textbf{Skin-SHARE} & \textbf{PBMC-TEA} & \textbf{PBMC-DOGMA} \\
        \midrule
        \multirow{2}{*}{ARI} & MoRE-GNN  & \textbf{0.892} & \textbf{0.570} & 0.691 & 0.247 & \textbf{0.657} & 0.359 \\
        & MOJITOO   & 0.868 & 0.545 & \textbf{0.774} & \textbf{0.441} & 0.654 & \textbf{0.472} \\
                             
        \midrule
        \multirow{2}{*}{NMI} & MoRE-GNN  & \textbf{0.881} & \textbf{0.699} & 0.723 & 0.458 & 0.741 & 0.487 \\
        & MOJITOO   & 0.861 & 0.672 & \textbf{0.782} & \textbf{0.632} & \textbf{0.751} & \textbf{0.558} \\
                             
        \bottomrule
    \end{tabular}}
\end{table}

MoRE-GNN improves clustering performance over MOJITOO on BM-CITE and LUNG-CITE, with an ARI of 0.892 and 0.570 and NMI of 0.881 and 0.699, respectively. These results indicate that when robust RNA–protein correlations exist, as is typical in CITE-seq data, the nonlinear, graph-based approach can better capture local neighbourhood structures.
On PBMC-Multiome and the tri-modal PBMC datasets, MoRE-GNN performs comparably to or slightly worse than MOJITOO, indicating that the model struggles to outperform linear methods in settings characterized by high dimensionality and substantial noise, as is common with ATAC-seq. Performance is poorest on Skin-SHARE, likely due to the presence of continuous cell-state 
gradients that hinder the formation of well-defined clusters.

\begin{figure}[h]
    \centering
        \includegraphics[width=0.65\linewidth]{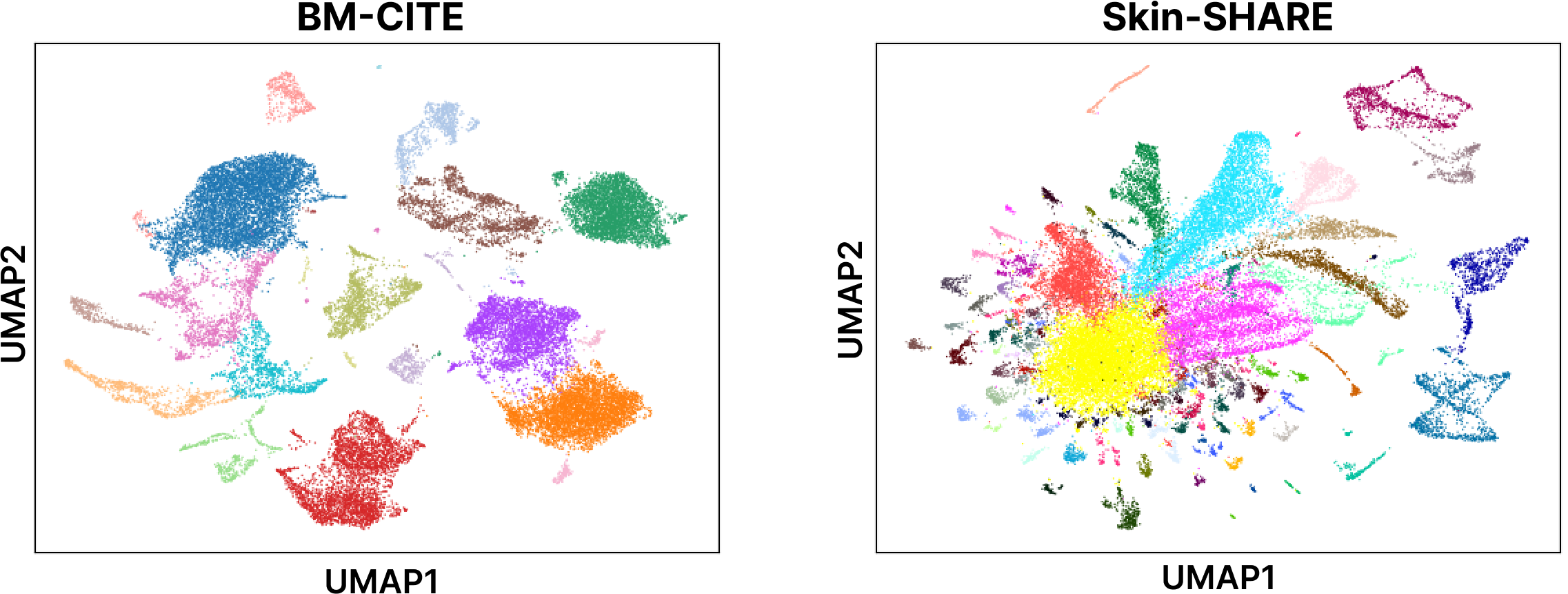}
    \caption{UMAP visualization of the latent representations produced by MoRE-GNN and clustered by Louvain algorithm. \textit{Left}: On BM-CITE, discrete immune cell populations are well-seperated into distinct clusters. \textit{Right}: On Skin-SHARE, however, the continuous spectrum of cell states leads to less defined cluster boundaries.}
    \label{fig:louvain-sp}
\end{figure}

To further probe why performance differs across datasets, we examined the latent space learned by MoRE-GNN by applying PCA and visualizing the first two dimensions. As shown in Figure \ref{fig:latent_sp}, datasets such as BM-CITE (and likewise the majority of remaining datasets in Appendix \ref{app:latent-space}) yield well-separated, triangle-shaped embeddings, while Skin-SHARE produces a diffuse structure. Figure~\ref{fig:louvain-sp} of downstream Louvain clustering gives further insights into the characteristics of the data types in each dataset (see Appendix \ref{app:louvain-discussion}). Clustering data from BM-CITE results in discrete partitions, while MoRE-GNN struggles to recover distinct cell communities from Skin-SHARE.

\begin{figure}[tb]
    \centering
        \includegraphics[width=0.6\linewidth]{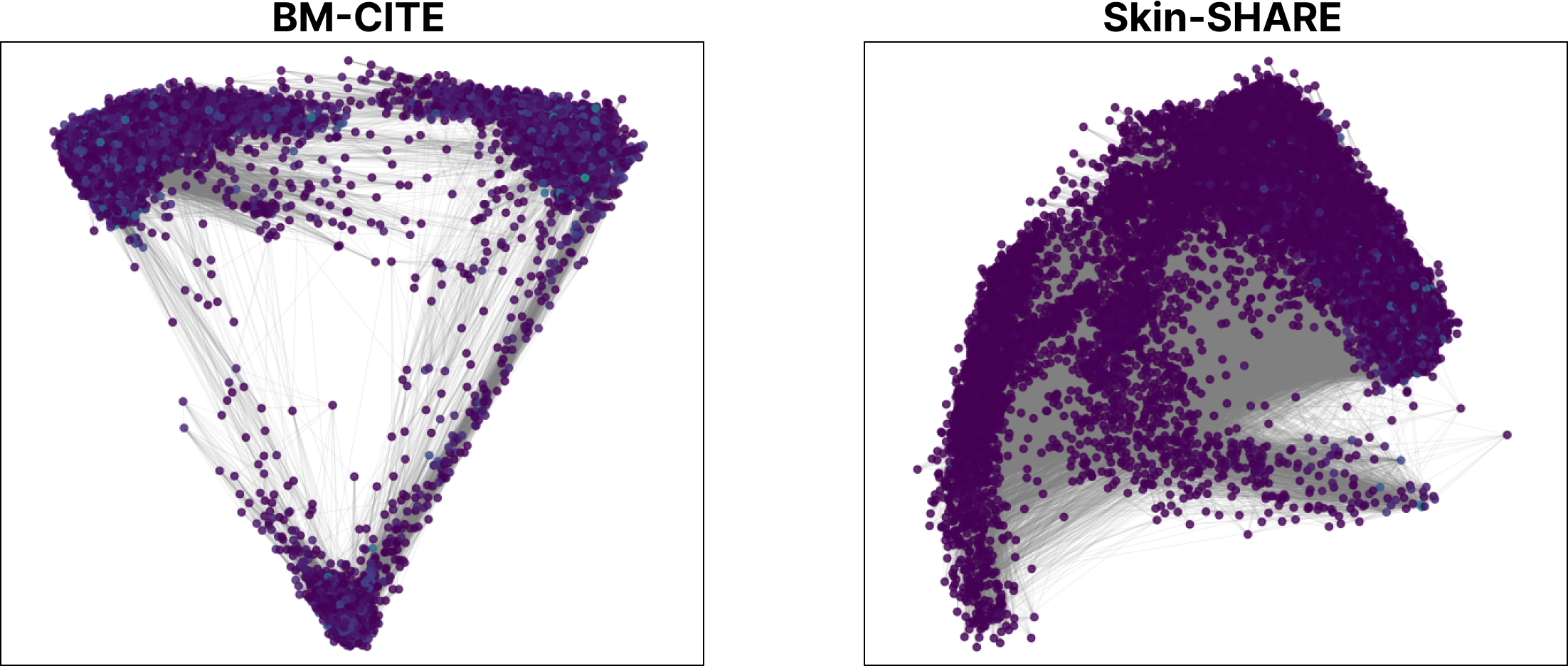}
    \caption{Two-dimensional PCA visualization of MoRE-GNN latent embeddings. Grey lines represent model-predicted edges, with node coloring reflecting degree (connectivity). \textit{Left}: BM-CITE dataset containing three major cell populations (T cells, B cells, and myeloid cells) exhibits clearly defined triangular clustering patterns in the latent space. \textit{Right}: Skin-SHARE dataset displays a more amorphous latent cloud structure due to the continuous gradient of cell states present in the biological system. The distinct geometric patterns reflect the underlying cellular composition and differentiation states within each dataset.}\label{fig:latent_sp} 
\end{figure}

\subsection{Cross-modality Feature Reconstruction}

Cross-modal prediction evaluates how well the latent embeddings retain information across modalities by predicting a target modality (e.g., RNA, accessibility, or protein abundance) directly from the embeddings \(z\). To this end, we use a lightweight multi layer perceptron (MLP) trained with mean squared error (MSE) loss. The predictor is intentionally left simple: a strong performance under this setting indicates that the latent embeddings themselves retain sufficient cross-modal information, rather than reflecting the capacity of a complex decoder. In other words, the simplicity of the MLP ensures that the evaluation probes the structure of the latent space, not the expressiveness of the prediction model. To evaluate the quality of cross-modal predictions, we compute Root Mean Squared Error (RMSE) \cite{willmott2005advantages_rmse} and Pearson Correlation Coefficient (PCC) \cite{cohen2009pearson}.

Table \ref{tab:cross_modality} summarizes MoRE-GNN’s performance when reconstructing one modality from the learned latent space, showing RMSE and PCC across different datasets. MoRE-GNN achieves notably lower RMSE and higher PCC when predicting ADT (surface protein) in BM-CITE compared to other modalities and datasets. This suggests that strong RNA–protein correlations in CITE-seq data help the model learn a latent representation that closely captures protein abundance. ATAC and Peaks predictions show higher RMSE and lower PCC, indicating that chromatin accessibility signals are less directly correlated with the latent features or require more specialized modeling strategies.

PBMC-Multiome shows a strong performance for RNA and Peaks. Skin-SHARE has consistently lower PCC scores. This finding aligns with the notion that continuous differentiation processes in Skin-SHARE and the complexity of open-chromatin data can be challenging for the proposed model and for the downstream cross-modality mapping. Nonetheless, MoRE-GNN still manages to capture some biologically relevant signal, as evidenced by PCC values that remain significantly above zero for all cases. Overall, these results highlight that MoRE-GNN is capable of reconstructing a target modality from the latent space, with the effectiveness of this undertaking depending on the intrinsic correlation between modalities and the biological complexity of the dataset.

\begin{table}[h]
\centering
\caption{Cross-modality prediction evaluation across modalities and datasets.}
\label{tab:cross_modality}
\renewcommand{\arraystretch}{1.1}
\resizebox{0.8\textwidth}{!}{%
\begin{tabular}{lcccccccc}
\hline
\multirow{2}{*}{\textbf{Dataset}} & \multicolumn{2}{c}{\textbf{RNA}} & \multicolumn{2}{c}{\textbf{ADT}} & \multicolumn{2}{c}{\textbf{ATAC}} & \multicolumn{2}{c}{\textbf{Peaks}} \\
\cmidrule(lr){2-3} \cmidrule(lr){4-5} \cmidrule(lr){6-7} \cmidrule(lr){8-9}
& RMSE & PCC & RMSE & PCC & RMSE & PCC & RMSE & PCC \\
\hline
BM-CITE & 0.790 & 0.405 & 0.307 & 0.952 & - & - & - & - \\
LUNG-CITE & 0.650 & 0.411 & 0.661 & 0.739 & - & - & - & - \\
PBMC-Multiome & 0.754 & 0.422 & - & - & - & - & 0.831 & 0.370 \\
Skin-SHARE & 0.761 & 0.303 & - & - & - & - & 0.761 & 0.303 \\
PBMC-TEA & 0.723 & 0.338 & 0.554 & 0.832 & 0.644 & 0.323 & - & - \\
PBMC-DOGMA & 0.707 & 0.406 & 0.825 & 0.562 & 0.814 & 0.458 & - & - \\
\hline
\end{tabular}%
}
\end{table}

\begin{figure}
    \centering
    \includegraphics[width=\linewidth]{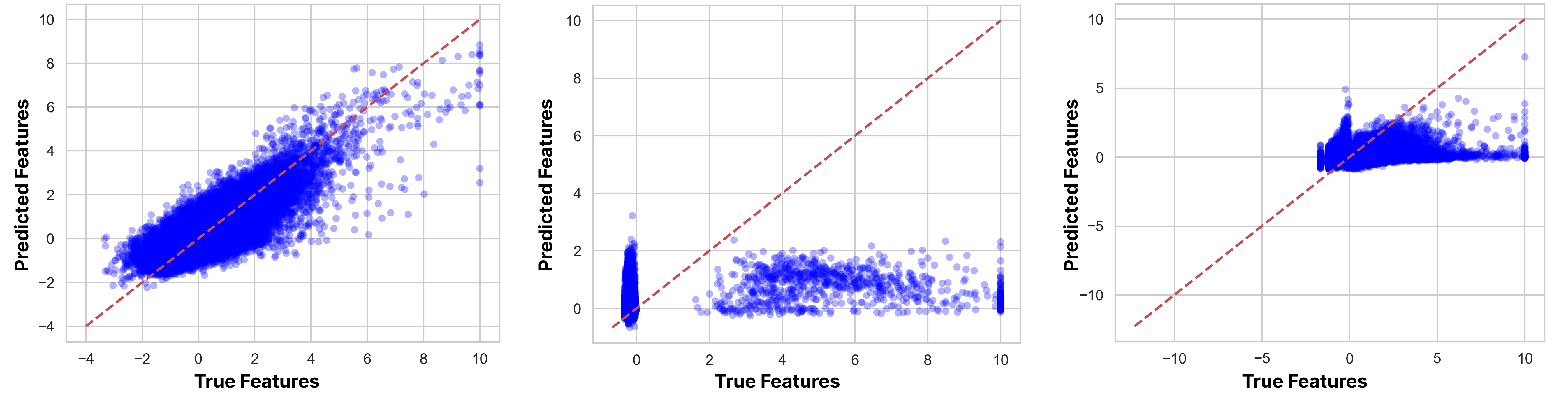}
    \caption{
    Cross-modality feature reconstruction from latent space on PBMC-TEA. Scatter plots show predicted vs. true feature values for (\textit{Left}) ADT, (\textit{Middle}) ATAC, and (\textit{Right}) RNA. The red dashed line indicates perfect prediction. ADT prediction exhibits strong concordance with minimal systematic bias, though variance increases for higher-abundance features. ATAC prediction is more challenging: predictions capture low-abundance signals but systematically underestimate high-intensity features. RNA prediction shows moderate correlation, but high variability and compressed dynamic range suggest difficulty in modeling its broader expression spectrum. These trends highlight that MoRE-GNN embeddings capture biologically meaningful cross-modal structure, but its performance is strongly modality-dependent.
    }
    \label{fig:cross-modal-pbmc-tea}
\end{figure}


Figure \ref{fig:cross-modal-pbmc-tea} illustrates modality-specific performance of cross-modal prediction on PBMC-TEA. In ADT prediction, values align closely with the diagonal, indicating strong accuracy and minimal bias, consistent with the high PCCs observed across datasets (Table \ref{tab:cross_modality}), such as PCC $=0.952$ on BM-CITE. This reflects the relatively low-dimensional, discrete nature of surface protein abundance, which is more directly learnable from latent embeddings.

In contrast, ATAC prediction shows systematic underestimation of high-intensity features, compressing the dynamic range and yielding moderate correlations (PCC $\approx 0.32$--0.46), consistent with the sparsity and noise of chromatin accessibility data. RNA prediction falls between these extremes: while correlations are moderate (PCC $\approx 0.30$--0.42), predictions cluster toward the mean, underrepresenting extreme values. This is likely caused by the high-dimensional and heterogeneous RNA expression space. Overall, MoRE-GNN captures biologically meaningful cross-modal structure, but predictive performance systematically depends on modality, with ADT consistently being the most predictable.

\section{Discussion}
Our experimental results reveal important insights about the complementary strengths of different multi-omics integration approaches. MOJITOO's reliance on a CCA-based framework inherently favours scenarios where cross-modal relationships are linear, which explains its superior performance on PBMC-Multiome and the noisy Skin-SHARE and PBMC-DOGMA datasets. Its method of aligning data using global linear projections proves robust in these challenging conditions. In contrast, our heterogeneous graph autoencoder is designed to capture more complex, nonlinear interactions through local neighbourhood structures, which yields a clear advantage in datasets like BM-CITE and LUNG-CITE where strong RNA–protein correlations provide a solid foundation for graph construction. The analysis of latent representations revealed a distinct triangular configuration in most datasets, likely reflecting the presence of two or three major cell lineages with transitional states forming the edges, confirming that MoRE-GNN can effectively disentangle discrete cell types from continuous biological processes.

However, this increased modeling flexibility comes at the cost of heightened sensitivity to hyperparameter settings, including learning rate, negative sampling strategies, and graph architecture choices, making MoRE-GNN more susceptible to performance degradation when faced with challenging data characteristics such as high dimensionality or extensive noise. The cross-modal prediction evaluation further demonstrates both the strengths and limitations of our approach: while ADT prediction achieved notably low RMSE and high PCC scores, RNA prediction revealed moderate accuracy with some variance and bias at higher expression levels. These findings suggest that while our model can outperform established methods under specific conditions, particularly when dealing with datasets exhibiting strong cross-modal correlations, its robustness across a wider range of biological contexts and data characteristics requires further refinement. Future work should focus on incorporating trajectory-aware clustering metrics and additional regularization techniques to improve the model's handling of complex and noisy modalities.

\section{Conclusion}
This work demonstrates that data-driven graph construction can significantly advance multi-omics integration by eliminating dependence on fixed biological priors while maintaining computational efficiency. The superior performance on datasets with strong cross-modal correlations suggests that our approach is particularly well-suited for applications where modality relationships are complex but structured. Moving forward, enhancing robustness across diverse biological contexts will be crucial for broader adoption in precision medicine and systems biology applications.

\bibliography{references}

\appendix
\section{Dimensionality Reduction.}
\label{app:pca_vs_no}

A common practice in single-cell data analysis is to apply dimensionality reduction (e.g., via PCA) prior to model training. However, for our graph autoencoder, we found that such preprocessing consistently degraded performance. Unlike linear methods, our model learns a shared latent space directly from raw features by reconstructing cell--cell edges, and does not rely on strong linear assumptions. Across datasets, using PCA reduced clustering quality, while training on normalized raw features yielded higher ARI and NMI.

\begin{figure}[H]
    \centering
    \begin{subfigure}{0.48\textwidth}
        \includegraphics[width=1\linewidth]{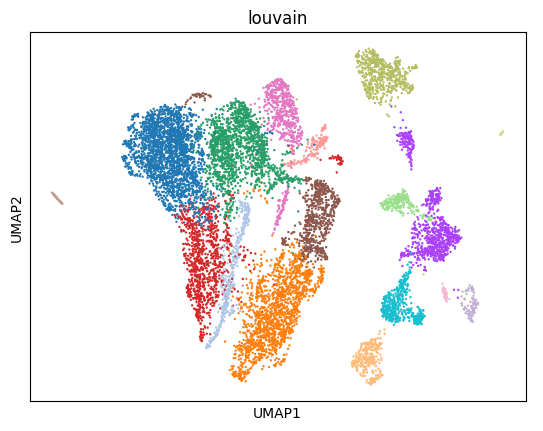}
        \caption{Visualized Louvain cluster assignments of LUNG-CITE (PCA). ARI = 0.489, NMI = 0.597.}
        \label{fig:louvain_lung_cite_pca_ed}
    \end{subfigure}
    \hfill
    \begin{subfigure}{0.48\textwidth}
        \includegraphics[width=1\linewidth]{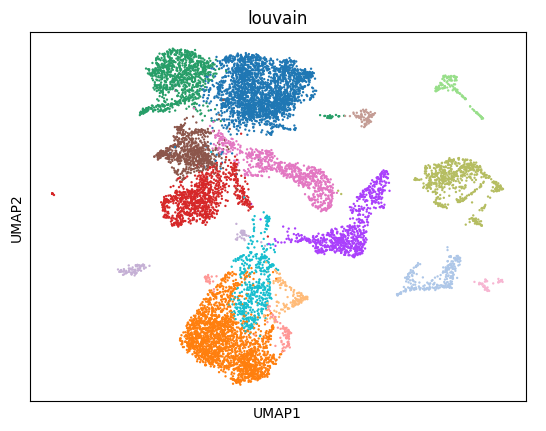}
        \caption{Visualized Louvain cluster assignments of LUNG-CITE (raw features). ARI = 0.545, NMI = 0.672.}
        \label{fig:lung_cite_louvain_raw}
    \end{subfigure}
    \caption{Effect of PCA preprocessing on clustering quality. Louvain cluster assignments for LUNG-CITE are shown using PCA rediced features (Left) and raw normalized features (Right). Clusters are more visually coherent without PCA, which is consistent with higher recorded ARI (0.545 vs. 0.489) and NMI (0.672 vs. 0.597).}
\end{figure}

To illustrate the impact of dimensionality reduction, we visualize Louvain cluster assignments on the LUNG-CITE dataset with and without PCA preprocessing (Figure \ref{fig:louvain_lung_cite_pca_ed}). Removing PCA yields more coherent and well-separated clusters, which is consistent with the quantitative evidence.

\section{Latent Representations} 
\label{app:latent-space}

\begin{figure}[tb]
    \centering
        \includegraphics[width=1.0\linewidth]{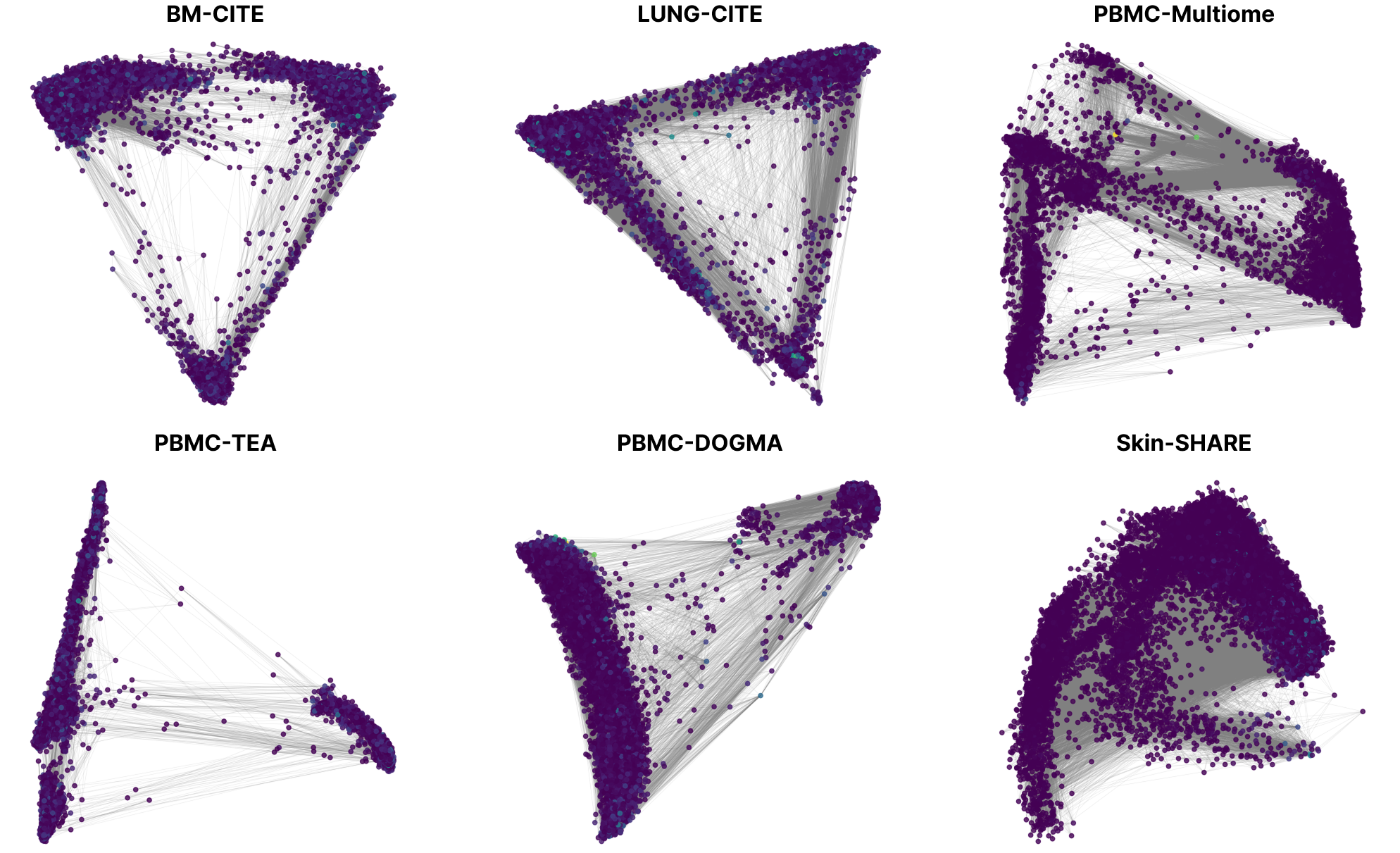}
    \caption{Latent embeddings learned by MoRE-GNN are projected into two dimensions using PCA. Grey lines represent edges predicted by the model, with node color intensity reflecting each node's connectivity. In many datasets (e.g., BM-CITE, PBMC-TEA) where the biological system contains three major cell populations (e.g, T cells, B cells and myeloid cells), the embeddings form triangular structures. The continuous gradient of cell states in Skin-SHARE dataset causes the latent cloud to appear more amorphous.}\label{fig:latent_grid_app} 
\end{figure}

The heterogeneous graph autoencoder produces \(d\)-dimensional embeddings \(z_i\) for each cell \(i\). These embeddings capture all of the multi-omic relationships in a single vector for each cell. We visualize the model's latent space in Figure \ref{fig:latent_grid_app} by finding two principal components of the \(d\)-dimensional latent vectors and projecting them into a 2-dimensional space. Each dot on the graph represents each cell's latent embedding \(z_i\) reduced to 2D using PCA.

Prior to training, for each modality, we compute a cosine-similarity matrix and for each cell, we mark its top-K most similar neighbors as positive edges. In Figure \ref{fig:latent_grid_app}, the grey edges represent pairs of cells \(i, j\) which form a positive edge in at least one modality. This results in the high-degree hubs of cells with many grey lines around them. 

In many of the datasets (BM-CITE, LUNG-CITE, PBMC-Multiome, PBMC-TEA, and PBMC-DOGMA) the biological system contains three major cell populations (e.g., T cells, B cells and myeloid cells). The visualized learned latent spaces for these datasets have triangular shapes which likely show each of these cell populations at one of the corners of the triangle. Transitional or mixed-phenotype cells lie along the edges between the corners. The three vertices of the triangle thus probably represent the three dominant lineages, and the edges capture intermediate states or cells that share features of two or more lineages. This reflects the true biological structure being captured by the graph autoencoder. The network edges drawn come directly from the data-driven kNN-style graphs that MoRE-GNN learns.

Skin tissue undergoes continuous differentiation rather than forming discrete clusters \cite{fuchs1990epidermal,cockburn2022gradual}. The continuous gradient of cell states causes the latent cloud to appear more amorphous rather than forming a clear triangular shape.

\section{Louvain Clustering}
\label{app:louvain-discussion}

\begin{figure}[tb]
    \centering
        \includegraphics[width=1.0\linewidth]{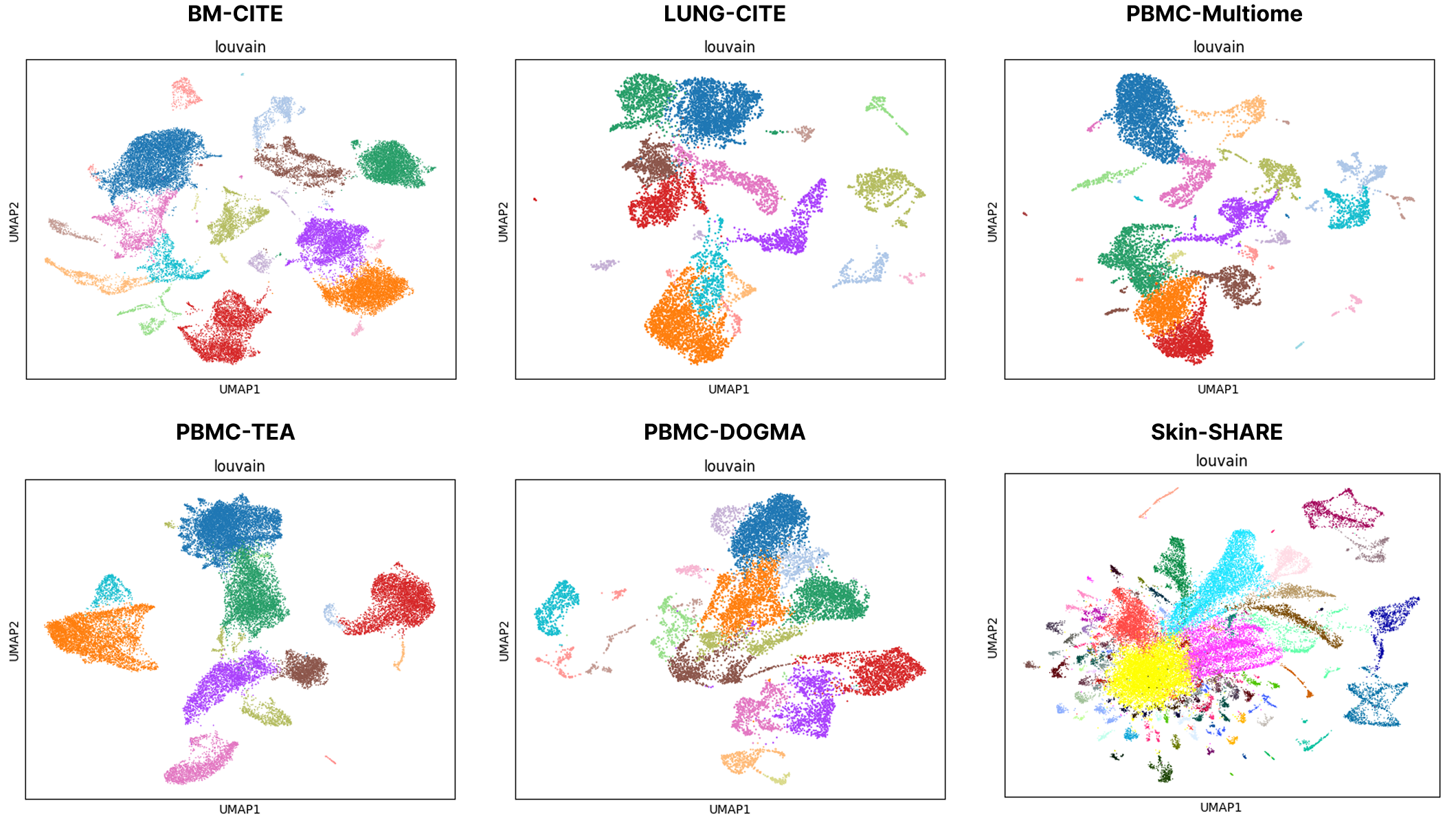}
    \caption{UMAP visualization of MoRE-GNN latent embeddings, colored by Louvain clusters. On datasets such as BM-CITE or PBMC-TEA, distinct cell populations are well-separated into discrete clusters, whereas continuous differentiation processes in Skin-SHARE lead to less sharply defined cluster boundaries.}
    \label{fig:louvain-grid_app}
\end{figure}

Figure \ref{fig:louvain-grid_app} shows the Louvain clustering results from the latent representations learned by MoRE-GNN across the six datasets. In BM-CITE and PBMC-TEA, where RNA–protein correlations tend to be stronger, the clusters appear well-separated. This outcome suggests that surface protein measurements can serve as robust anchors for integration, making it easier for the model to distinguish different immune cell populations. By contrast, LUNG-CITE and PBMC-Multiome show more overlapping clusters, likely reflecting the additional complexity of chromatin variability or weaker cross-modal correlations. Although MoRE-GNN still identifies broad cell-type groupings, epigenetic noise and subtle biological differences make it challenging to achieve the same level of clarity seen in CITE-seq data.

PBMC-DOGMA and Skin-SHARE present further difficulties due to their multimodal or continuous biological processes. In PBMC-DOGMA, the model successfully resolves major cell groups, but the presence of three integrated modalities introduces extra complexity. Skin-SHARE, in particular, captures a gradual differentiation process in skin cells, which complicates discrete clustering methods. Consequently, while MoRE-GNN integrates the data into a biologically meaningful latent space, it struggles to produce neat clusters for tissues characterized by more fluid cell-state transitions. Overall, these observations indicate that MoRE-GNN excels when strong modality alignments exist (such as RNA–protein pairs), but continuous gradients or high levels of noise can diminish its ability to form distinct clusters. 

\section{RNA cross-modality prediction task on BM-CITE}
\label{sc:rna-cross-mod}

\begin{figure}[h]
    \centering
    \includegraphics[width=0.4\linewidth]{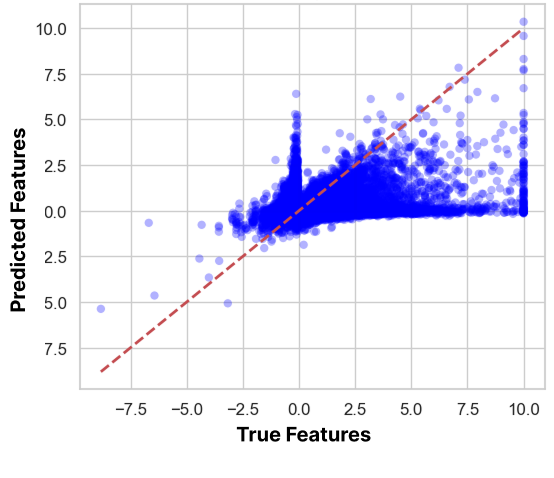}
    \caption{Evaluation of RNA prediction task from latent space trained on BM-CITE. True vs. predicted values. The red dashed line represents the perfect prediction. Points are moderately spread from the line, indicating some level of error.}
    \label{fig:rna-cross-modality-bm-cite}
\end{figure}

Figure \ref{fig:rna-cross-modality-bm-cite} presents a case study of modality-specific reconstruction, evaluating RNA prediction from the latent space learned on BM-CITE. The scatter plot presented in Figure \ref{fig:rna-cross-modality-bm-cite} shows that the predicted RNA values are moderately dispersed around the ideal prediction line (shown as a red dashed line). This indicates that, although the latent representation captures a substantial part of the true expression variance, there remains considerable variability, resulting in moderate RMSE and PCC scores. The prediction errors are centred around zero, yet a slight asymmetry suggests that the model tends to underestimate the RNA expression in some cases. Most of (true, predicted) pairs cluster along the diagonal, which indicates an overall absence of systematic bias; however, the high variance points to limitations in capturing extreme expression values. There is an increasing error with higher mean values, implying that the model’s predictive performance deteriorates for cells with higher RNA expression levels.

\end{document}